\documentclass[10pt,twocolumn,letterpaper]{article}
\usepackage[pagenumbers]{cvpr}

\usepackage[T1]{fontenc}
\usepackage{times}
\usepackage{graphicx}
\usepackage{cuted}
\usepackage{amsmath,amssymb}
\usepackage{booktabs}
\usepackage{balance}
\usepackage{microtype}
\usepackage{xcolor}
\definecolor{cvprblue}{rgb}{0.21,0.49,0.74}
\usepackage[breaklinks=true,colorlinks=true,allcolors=cvprblue]{hyperref}
\hypersetup{pdfborder={0 0 0}}

\begin{document}

\title{PGN: Design and Implementation of a Vision-Language Navigation System Based on Pangu Multimodal Foundation Model}

\author{
    Li Xian, Mingxi Li, Yizheng Wang, Yiming Shen, Qi Chen, Zhuoling Xiao\textsuperscript{*}\\
    School of Information and Communication Engineering\\
    University of Electronic Science and Technology of China\\
    {\small \textsuperscript{*}Corresponding author: \texttt{zhuolingxiao@uestc.edu.cn}}
}

\maketitle

\begin{abstract}
Vision-Language Navigation (VLN) requires an embodied agent to interpret a natural-language instruction and predict actions from temporally ordered visual observations. Adapting a multimodal large language model to VLN requires visual-language alignment, compact temporal inputs, action-space grounding, and stable training on the target hardware. This technical report presents PGN (Pangu Navigator), an offline VLN action-prediction system built on OpenPangu-7B. Training proceeds in two stages. First, PGMM aligns a frozen EVA-ViT-G/14 vision encoder with the frozen language backbone by training a Q-Former and a two-layer MLP projector. Second, PGN adapts the aligned model to expert navigation trajectories using five-observation windows, epoch-dependent temporal sampling, and a reasoning-then-action output format; this stage freezes the aligned visual pathway and updates three structural-token embeddings and LoRA adapters. The implementation combines mixed-precision computation, selective FP32 computation, and DeepSpeed ZeRO-2 on eight Ascend 910B NPUs. Under teacher-forced, open-loop evaluation on 500 held-out expert trajectories, V9 reports a 62.29\% Normalized Action Match (NAM) and a 100.00\% Non-empty Rate (NER). These metrics quantify offline expert-action alignment rather than closed-loop navigation success; evaluating error accumulation, path efficiency, and goal completion remains future work.
\end{abstract}

\section{Introduction}

Vision-Language Navigation (VLN) connects visual perception, language understanding, spatial reasoning, and action generation. Given an instruction such as ``walk past the kitchen and stop near the vase,'' an agent must interpret landmarks, use recent observations as temporal context, and select a sequence of low-level actions. Early VLN benchmarks use discrete navigation graphs \cite{anderson2018vision}, while VLN-CE extends the task to continuous environments with forward and turning actions \cite{krantz2020beyond}.

Multimodal large language models (MLLMs) offer broad visual-language representations and flexible text generation \cite{caffagni2024revolution,li2023blip2}. However, adapting a general MLLM to continuous navigation requires aligning visual features with the language model, representing temporal observations compactly, grounding generated responses in a small action space, and maintaining an efficient training stack. These constraints are particularly relevant when a 7B-scale model is trained on Ascend NPUs, where unsupported or unstable half-precision operations may require selective full-precision computation.

This report describes PGN, an implementation that adapts OpenPangu-7B \cite{chen2025panguembedded} to offline VLN action prediction. It addresses these requirements through two-stage visual-language alignment and navigation adaptation. The current study evaluates offline expert-action alignment under teacher-forced, open-loop observation histories rather than closed-loop navigation success. The contributions are:
\begin{itemize}
    \item A two-stage system that first aligns visual and language representations through PGMM and then adapts the aligned model to expert navigation trajectories with temporal observations and LoRA.
    \item A structured navigation input and output format combining five-frame observation windows, temporal boundary tokens, reasoning text, and a canonical four-action space.
    \item An Ascend-oriented training implementation and a teacher-forced, open-loop evaluation across model iterations, together with a discussion of what offline action matching can and cannot establish.
\end{itemize}

\section{Related Work}

\paragraph{Vision-language navigation.}
R2R introduced language-guided navigation in real scanned environments \cite{anderson2018vision}, and RxR expanded the task with multilingual, densely grounded instructions \cite{ku2020room}. VLN-CE extended navigation from predefined graph nodes to continuous environments \cite{krantz2020beyond}. History-aware transformers \cite{chen2022history} and in-domain pretraining \cite{guhur2021airbert} improve temporal and cross-modal representations. More recent work uses large language models for explicit navigation reasoning \cite{zhou2023navgpt} and navigation instruction tuning \cite{huang2024navillm}. These studies motivate PGN's use of temporally ordered observations and instruction-conditioned action prediction.

\paragraph{Multimodal foundation models.}
BLIP-2 uses a Q-Former to connect a frozen vision encoder and a frozen language model \cite{li2023blip2}. LLaVA demonstrates the effectiveness of visual instruction tuning \cite{liu2024visual}, while DeepSeek-VL2 explores sparse multimodal architectures \cite{wu2024deepseekvl2}. PGN follows this frozen-backbone alignment pattern while using OpenPangu-7B as the language backbone and an Ascend NPU training stack.

\paragraph{Reasoning and parameter-efficient adaptation.}
Chain-of-thought prompting elicits intermediate reasoning text before a final response \cite{wei2022chain}, and LoRA enables task adaptation through low-rank updates rather than full-model fine-tuning \cite{hu2021lora}. In PGN, reasoning text precedes the final canonical action, while LoRA adapters limit updates to the language backbone.

\section{PGN Method}

\begin{figure*}[t]
    \centering
    \includegraphics[width=0.64\textwidth]{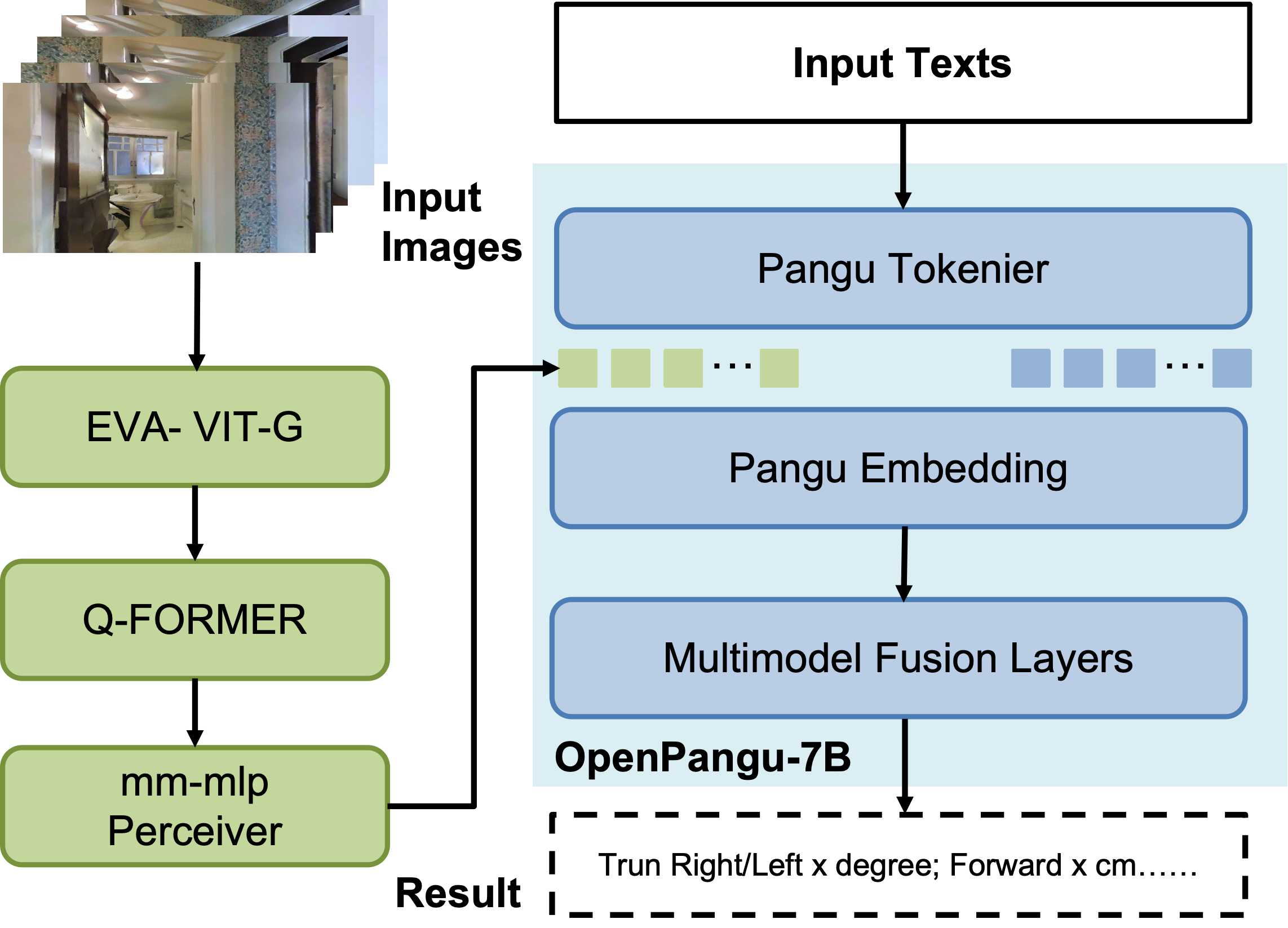}
    \caption{PGN system topology. Five temporally ordered observations pass through the shared EVA-ViT-G, Q-Former, and multimodal MLP pathway. Their visual tokens are assembled with the tokenized navigation instruction in the OpenPangu-7B embedding space and processed by the multimodal fusion layers.}
    \label{fig:pgn_overall}
\end{figure*}

\subsection{Navigation Formulation}

At time step $t$, PGN receives a navigation instruction $I$ and an ordered window containing four historical observations and the current observation:
\begin{equation}
    O_t = (o_{t-4s}, o_{t-3s}, o_{t-2s}, o_{t-s}, o_t),
\end{equation}
where $o_{\tau}$ denotes the observation at time $\tau$, and $s$ is the temporal sampling stride. The window length remains fixed at five observations, while $s$ controls the temporal spacing between them. The observations retain their chronological order in the serialized multimodal input, and special structural tokens mark image starts, image ends, and frame boundaries. The action space is
\begin{equation}
    \mathcal{A}=\{\texttt{forward},\texttt{left},\texttt{right},\texttt{stop}\},
\end{equation}
where the four symbols denote semantic action classes. In the collected trajectories, forward motion is 0.25~m and each turn is 15 degrees. The model generates reasoning text $r_t$ before the action $a_t$:
\begin{equation}
P(r_t,a_t\mid I,O_t)=P(r_t\mid I,O_t)P(a_t\mid r_t,I,O_t).
\end{equation}
This factorization specifies the output format; the present experiments do not separately measure reasoning quality.

\subsection{Model Components}

Figure~\ref{fig:pgn_overall} summarizes the data flow. PGN denotes the complete navigation action-generation system, whereas OpenPangu-7B serves as its language backbone. For each $224\times224$ observation, EVA-ViT-G/14 first extracts patch-level visual features. A Q-Former then uses 32 learnable query tokens and cross-attention to produce a fixed-size visual representation. A two-layer MLP projector maps each 768-dimensional query representation to the 4096-dimensional OpenPangu-7B embedding space, making the projected visual tokens dimensionally compatible with the language backbone. The component is labeled ``mm-mlp Perceiver'' in Figure~\ref{fig:pgn_overall}; we use the term ``two-layer MLP projector'' in the text.

The five observations are processed by this shared visual-alignment pathway, and their projected tokens remain temporally ordered. A navigation-assistant identity prompt, the instruction, and the ordered visual tokens form a single autoregressive context. Three learned structural tokens delimit image starts, image ends, and frame boundaries. OpenPangu-7B then generates reasoning text followed by a canonical action, while LoRA adapters provide navigation-specific updates without full-model fine-tuning.

\section{Data and Training}

PGN is trained in two stages. Stage~1 aligns visual and language representations using paired images and captions, and Stage~2 uses expert navigation trajectories to adapt the aligned model for supervised next-action generation. This division separates visual conditioning from navigation-task adaptation.

\subsection{Visual-Language Alignment}

Stage~1 trains PGMM on image-caption pairs from Microsoft COCO \cite{lin2014coco} and Visual Genome \cite{krishna2017visual}. As shown in Figure~\ref{fig:training}(a), the EVA-ViT-G/14 vision encoder and OpenPangu-7B language backbone remain frozen, while the Q-Former and two-layer MLP projector are optimized for image-conditioned text generation. This stage prepares visual inputs for the language backbone before navigation adaptation.

\subsection{Expert Trajectory Collection}

\begin{figure}[t]
    \centering
    \includegraphics[width=\linewidth]{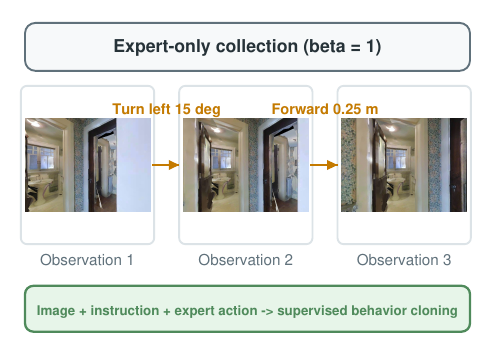}
    \caption{Example of expert-only trajectory collection. The data pipeline uses $\beta=1$, so every stored action label is produced by the expert policy and is used for supervised behavior cloning.}
    \label{fig:expert_data}
\end{figure}

The navigation dataset was collected through a DAgger-compatible simulation pipeline. At collection iteration $i$, the policy mixture is
\begin{equation}
    \pi_{\mathrm{active}}=\beta^i\pi^*+(1-\beta^i)\hat{\pi},
\end{equation}
where $\pi^*$ is the expert policy and $\hat{\pi}$ is the learned policy. The recorded configuration fixes $\beta=1$, so $\beta^i=1$ and the active policy is always the expert. Consequently, the procedure is expert-only data collection followed by supervised behavior cloning, rather than iterative on-policy DAgger training.

After cleaning, the development record contains 19,699 navigation trajectories and 1,567,487 observation frames. Each trajectory pairs a natural-language instruction and a sequence of observations with the expert action at each step. Figure~\ref{fig:expert_data} illustrates a short segment.

\subsection{Temporal and Action-Aware Sampling}

During Stage~2, five-observation windows are sampled from each trajectory. To vary the temporal offset while keeping the window length fixed, the starting index changes with epoch $e$:
\begin{equation}
    i_{\mathrm{start}}=e\bmod s.
\end{equation}
The recorded Stage~2 sampling curriculum uses $(s,p)=(4,0.1)$, $(6,0.6)$, and $(8,0.2)$ in three successive phases. Here, $s$ is the temporal stride defined in Section~3.1, and $p$ is the probability that the action-aware sampler replaces a forward-dominated selection with an available turning or stopping sample. Rotating the starting offset and rebalancing action selection are separate operations.

\subsection{Optimization and Ascend Implementation}

\begin{figure*}[t]
    \centering
    \begin{minipage}[t]{0.49\textwidth}
        \centering
        \includegraphics[width=\linewidth]{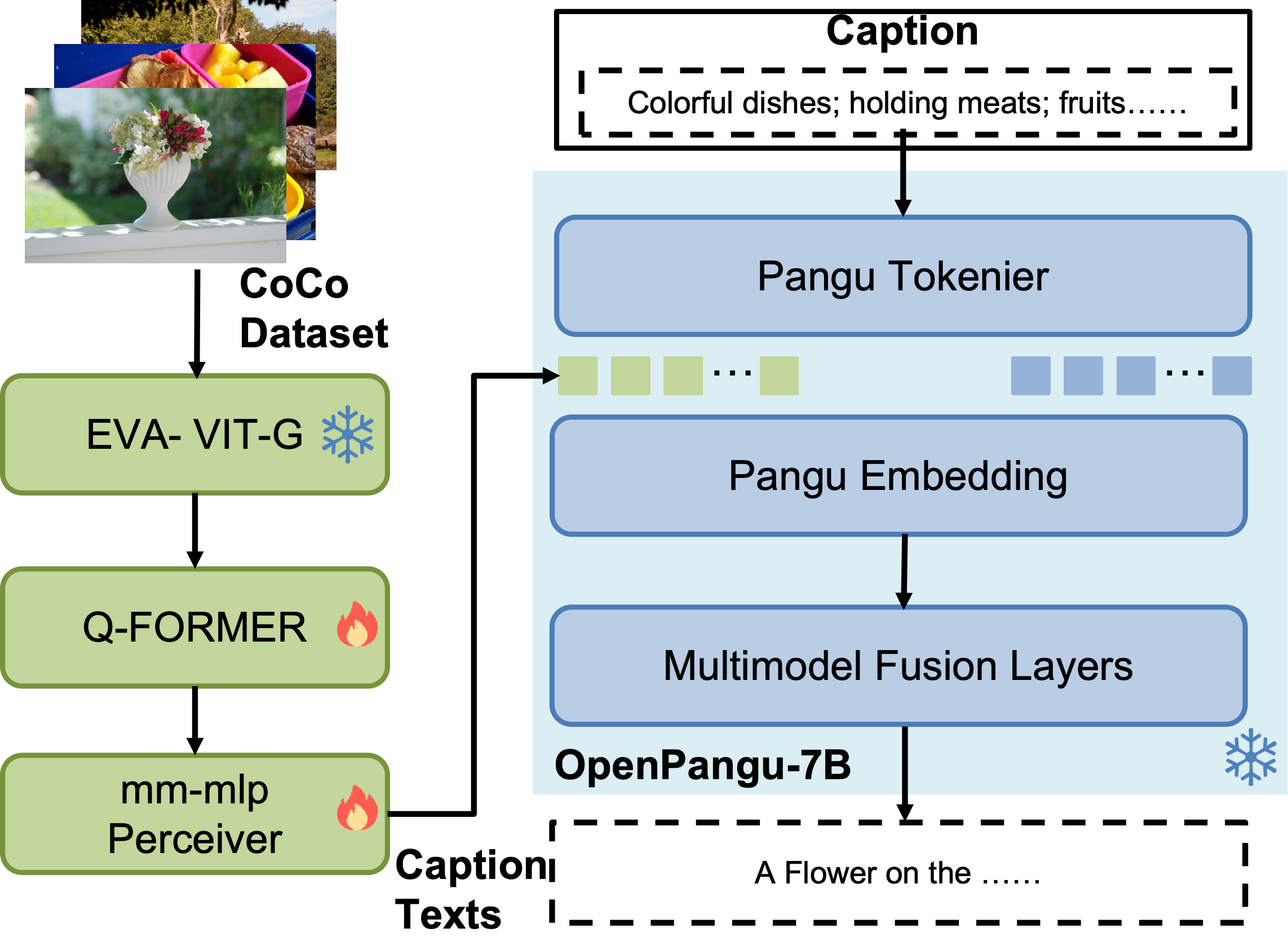}
        \vspace{-2pt}

        \small (a) Stage 1: multimodal alignment pretraining
    \end{minipage}
    \hfill
    \begin{minipage}[t]{0.49\textwidth}
        \centering
        \includegraphics[width=\linewidth]{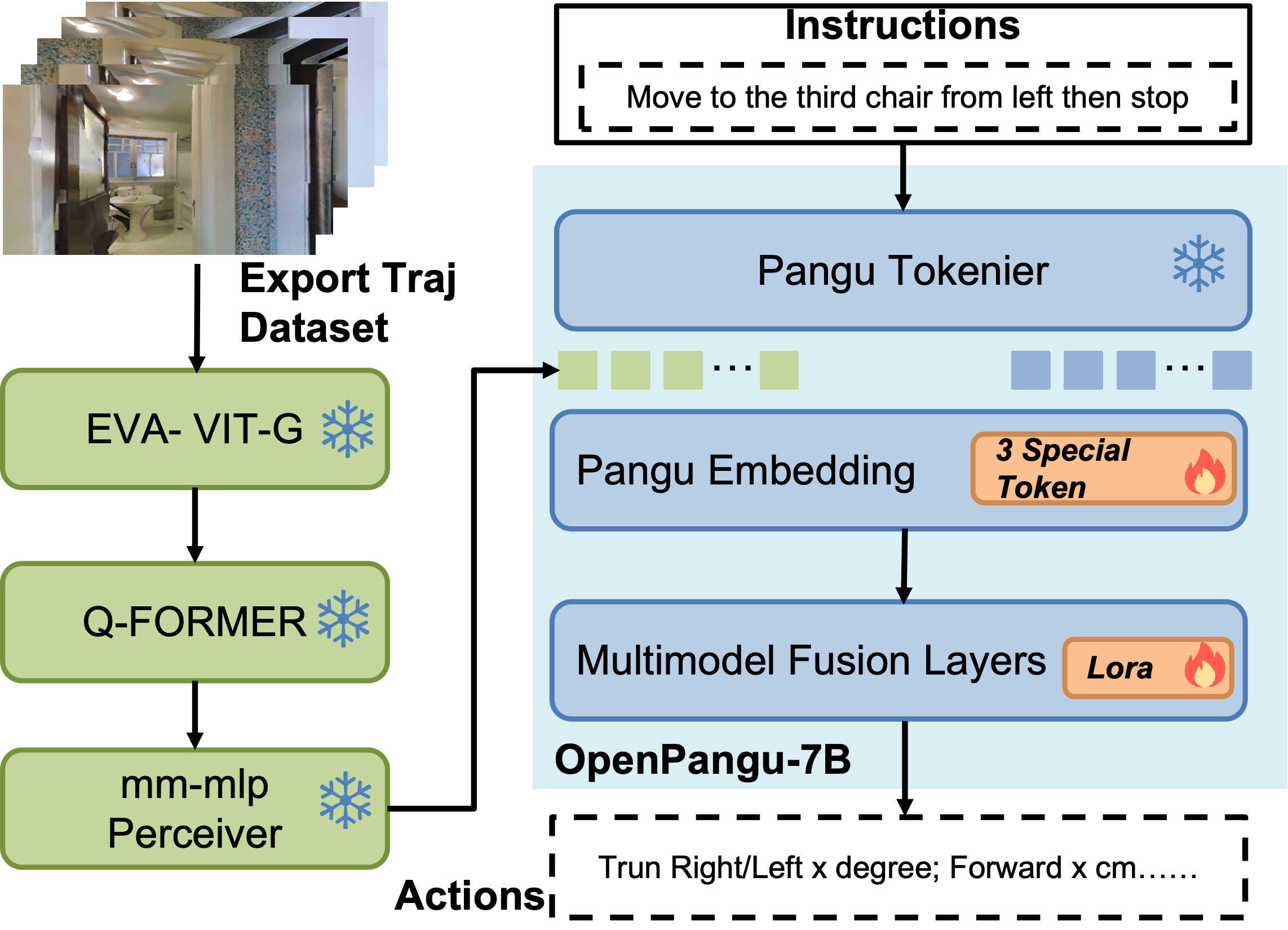}
        \vspace{-2pt}

        \small (b) Stage 2: navigation adaptation
    \end{minipage}
    \caption{Two-stage training architecture. Stage~1 freezes EVA-ViT-G and OpenPangu-7B while training the Q-Former and multimodal MLP projector on image-text pairs. Stage~2 freezes the aligned visual pathway and the base language model, while training three structural-token embeddings and LoRA adapters on expert navigation trajectories. Snowflakes and flames denote frozen and trainable components, respectively.}
    \label{fig:training}
\end{figure*}

Figure~\ref{fig:training} summarizes the parameter updates. Stage~1 updates the Q-Former and two-layer MLP projector while keeping both backbones frozen. In Stage~2, the aligned EVA-ViT-G/14, Q-Former, projector, tokenizer, and base OpenPangu-7B parameters remain frozen. Navigation-specific learning is confined to the three structural-token embeddings and the LoRA adapters injected into the language model.

The training stack uses eight Ascend 910B NPUs, mixed-precision computation, and DeepSpeed ZeRO-2. When FP16 attention-mask operations produced numerical overflow, only the affected computation was performed in FP32. FP32 was therefore used selectively for numerical stability rather than as the global training precision.

\section{Offline Evaluation}

\subsection{Protocol and Metrics}

\begin{figure}[t]
    \centering
    \includegraphics[width=0.98\linewidth]{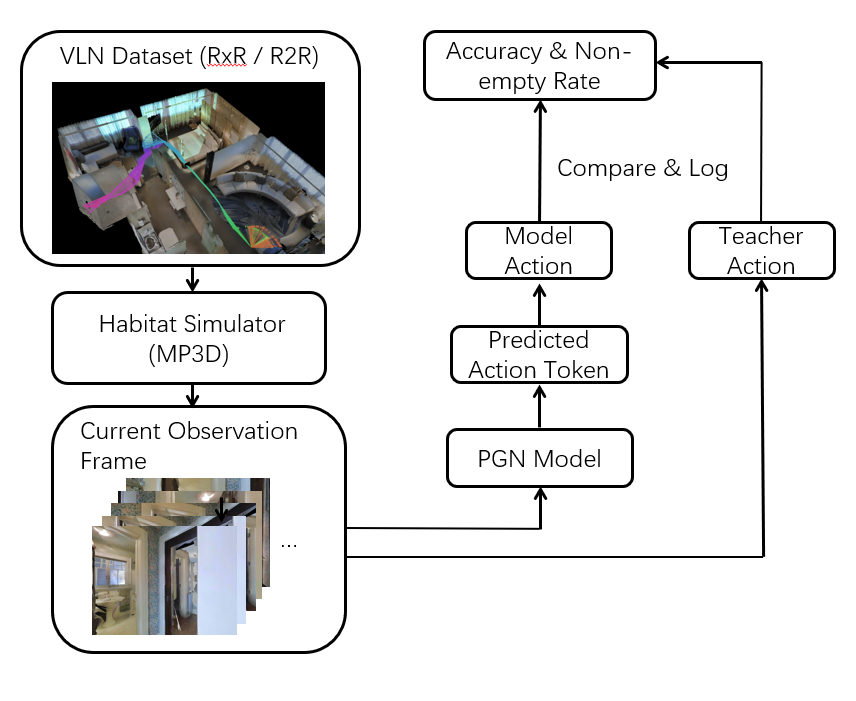}
    \caption{Offline expert-trajectory alignment evaluation. The held-out trajectory supplies the observation history and teacher action. PGN predicts an action token, which is compared with the teacher action to compute and log alignment metrics. The prediction is not executed to obtain the next observation.}
    \label{fig:offline_eval}
\end{figure}

The validation split contains 500 held-out expert trajectories. Evaluation uses stride $s=1$ and supplies the model with a fixed five-observation window: four historical observations and the current observation from the expert trajectory. Figure~\ref{fig:offline_eval} retains the evaluation flow from the prior manuscript. Although the Habitat simulator is shown as the source of the recorded trajectory, the model prediction does not change the next evaluation observation. The protocol therefore measures teacher-forced local action agreement without closed-loop error accumulation.

We report three metrics:
\begin{itemize}
    \item \textbf{Exact-string Match (EM):} the generated response exactly equals the canonical action string.
    \item \textbf{Normalized Action Match (NAM):} an action is extracted from the response after normalizing forward, left, right, and stop expressions, then compared with the expert action.
    \item \textbf{Non-empty Rate (NER):} the fraction of action predictions producing any non-empty response.
\end{itemize}
For V6, the retained raw file contains 40,000 action predictions over the 500 trajectories and reproduces EM $=11.61\%$, NAM $=23.18\%$, and NER $=99.17\%$.

\clearpage
\subsection{Results Across Iterations}

\begin{strip}
    \centering
    \refstepcounter{table}
    \label{tab:results}
    \parbox{\textwidth}{\small Table~\thetable. Original offline evaluation data reported across PGN iterations. Strict, Fuzzy, and Non-blank correspond to EM, NAM, and NER in the revised terminology. Metric values are percentages; the ten rightmost columns are raw counts of held-out trajectories in per-trajectory Fuzzy/NAM intervals.}

    \vspace{6pt}
    \setlength{\tabcolsep}{3.1pt}
    \scriptsize
    \begin{tabular}{lccc*{10}{r}}
        \toprule
        & \multicolumn{3}{c}{Action-level metrics (\%)} & \multicolumn{10}{c}{Per-trajectory Fuzzy/NAM interval counts} \\
        \cmidrule(lr){2-4}\cmidrule(lr){5-14}
        Version & Strict & Fuzzy & Non-blank & 0--10 & 10--20 & 20--30 & 30--40 & 40--50 & 50--60 & 60--70 & 70--80 & 80--90 & 90--100 \\
        \midrule
        V3 & 8.40 & 8.40 & 12.93 & 39 & 8 & 7 & 0 & 1 & 1 & 0 & 0 & 0 & 0 \\
        V5 & 25.40 & 25.40 & 33.98 & 41 & 115 & 160 & 113 & 40 & 18 & 10 & 2 & 1 & 0 \\
        V6 & 11.61 & 23.18 & 99.17 & 80 & 175 & 118 & 43 & 31 & 26 & 16 & 8 & 3 & 0 \\
        V8 & 0.00 & 62.13 & 100.00 & 5 & 1 & 15 & 26 & 51 & 141 & 147 & 85 & 26 & 3 \\
        V9 & 0.00 & 62.29 & 100.00 & 5 & 1 & 13 & 25 & 53 & 133 & 160 & 84 & 23 & 3 \\
        \bottomrule
    \end{tabular}

    \vspace{3pt}
    \parbox{0.98\textwidth}{\scriptsize V6 is reproducible from retained per-action predictions; V8 and V9 are retained aggregate summaries. The original V3 interval counts sum to 56 rather than 500, so V3 is preserved here for source fidelity but excluded from Figure~\ref{fig:performance}(b).}

    \vspace{18pt}

    \includegraphics[width=0.98\textwidth]{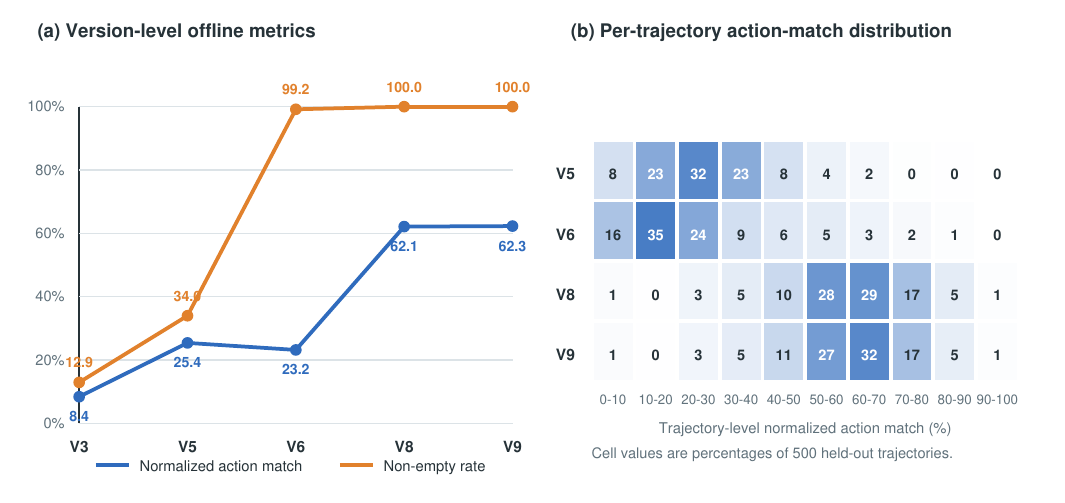}
    \refstepcounter{figure}
    \label{fig:performance}
    \parbox{\textwidth}{\small Figure~\thefigure. Visualization of the raw results in Table~\ref{tab:results}. (a) Fuzzy and Non-blank are shown using the revised NAM and NER terminology. (b) The raw interval counts for versions with internally consistent 500-trajectory totals are converted to percentages and displayed as a heatmap. V3 is omitted from the heatmap because its recorded counts sum to 56.}
\end{strip}

Table~\ref{tab:results} preserves the original metric values and interval counts, while Figure~\ref{fig:performance} visualizes the comparable trends and distributions. From V5 to V6, NER rises from 33.98\% to 99.17\%, while NAM changes from 25.40\% to 23.18\%. V8 and V9 report NAM values of 62.13\% and 62.29\%, respectively, with NER of 100.00\%. Their EM is zero because these versions generate reasoning text before the action; whole-response exact matching is therefore not comparable with versions that emit only an action phrase.

The per-trajectory distribution also moves from the 10--30\% region for V5/V6 toward the 50--80\% region for V8/V9. Reasoning-then-action output, identity prompting, temporal sampling, and action-aware sampling were introduced within the same development cycle. The available results therefore document the combined version change but do not isolate the causal contribution of each component.

\subsection{Evaluation Limitations}

The teacher-forced protocol measures action agreement under expert observation histories. It does not measure whether the model can recover from an incorrect action, stop within a goal radius, or follow an efficient path. Accordingly, the reported numbers are not Success Rate, SPL, navigation error, or nDTW. Closed-loop Habitat evaluation is required before making claims about navigation success.

In addition, per-action logs were retained for V6 but not for the later aggregate summaries. We therefore do not report action-wise error percentages for V8 or V9. The aggregate trend is useful as a development record, but it should not be interpreted as an ablation study or a comparison with published VLN systems.

\section{Conclusion}

This report presented PGN, an OpenPangu-7B-based system for offline VLN action prediction. PGN combines Q-Former-based visual alignment, a two-stage training pipeline, five-observation temporal inputs, LoRA adaptation, and an Ascend-oriented distributed training stack. V8 and V9 report NAM values of 62.13\% and 62.29\%, respectively, with NER of 100.00\% under teacher-forced, open-loop evaluation. These results support the feasibility of adapting OpenPangu-7B to offline expert-action alignment, but they do not establish closed-loop navigation success. Future evaluation should execute predicted actions in the simulator and report goal completion, path efficiency, trajectory similarity, and recovery from accumulated errors.

\balance
{\footnotesize
\setlength{\bibsep}{1pt plus 0.3ex}
\bibliographystyle{ieeenat_fullname}
\bibliography{references}
}

\end{document}